\documentclass[sigconf, nonacm]{acmart}
\usepackage{subcaption}
\usepackage{multirow}
\usepackage{colortbl}
\usepackage{pifont}
\usepackage{xspace}
\usepackage{xcolor}
\usepackage{textcomp}
\usepackage{wrapfig}
\usepackage{float}
\usepackage[most]{tcolorbox}
\usepackage{longtable}
\usepackage{dashrule}
\usepackage{makecell}
\usepackage{threeparttable}
\usepackage{enumitem}
\usepackage[normalem]{ulem}
\usepackage{tabularx}
\usepackage{mathtools}
\usepackage{amsthm}
\usepackage[capitalize,noabbrev]{cleveref}

\newcommand{\sandbox}{\textrm{LAST-Box}\xspace}
\newcommand{\method}{\textrm{LAST}\xspace}
\newcommand{\proposedmodel}{\textrm{LAST-7B}\xspace}

\newtcolorbox[auto counter, number freestyle={\noexpand\arabic{\tcbcounter}}]{definedbox}[2][]{%
    enhanced,
    colback=black!5!white,
    colframe=black!75!white,
    title=#2,
    #1
}

\newtcolorbox{formalizedprompt}[1][]{%
  enhanced,
  breakable,
  colback=gray!5,
  colframe=gray!80,
  fonttitle=\bfseries,
  title=Formalize2Tabular\_prompt,
  coltitle=black,
  sharp corners,
  boxrule=0.8pt,
  arc=2mm,
  left=4pt,
  right=4pt,
  top=6pt,
  bottom=6pt,
  width=\textwidth,
  #1
}

\begin{document}

\title{LAST: Leveraging Tools as Hints to Enhance Spatial Reasoning for Multimodal Large Language Models}

\author{Shi-Yu Tian}
\affiliation{%
  \institution{National Key Laboratory for Novel Software Technology, Nanjing University}
  \country{China}
}

\author{Zhi Zhou}
\affiliation{%
  \institution{National Key Laboratory for Novel Software Technology, Nanjing University}
  \country{China}
}

\author{Kun-Yang Yu}
\affiliation{%
  \institution{National Key Laboratory for Novel Software Technology, Nanjing University}
  \country{China}
}

\author{Ming Yang}
\affiliation{%
  \institution{National Key Laboratory for Novel Software Technology, Nanjing University}
  \country{China}
}

\author{Yang Chen}
\affiliation{%
  \institution{National Key Laboratory for Novel Software Technology, Nanjing University}
  \country{China}
}

\author{Ziqiao Shang}
\affiliation{%
  \institution{National Key Laboratory for Novel Software Technology, Nanjing University}
  \country{China}
}

\author{Lan-Zhe Guo}
\affiliation{%
  \institution{National Key Laboratory for Novel Software Technology, Nanjing University}
  \country{China}
}

\author{Yu-Feng Li}
\affiliation{%
  \institution{National Key Laboratory for Novel Software Technology, Nanjing University}
  \country{China}
}

\renewcommand{\shortauthors}{Tian et al.}

\begin{abstract}
Spatial reasoning is a cornerstone capability for intelligent systems to perceive and interact with the physical world. However, multimodal large language models (MLLMs) frequently suffer from hallucinations and imprecision when parsing complex geometric layouts.
As data-driven scaling struggles to internalize structured geometric priors and spatial constraints, integrating mature, specialized vision models presents a compelling alternative.
Despite its promise, applying this paradigm to spatial reasoning is hindered by two key challenges:
the difficulty of invoking heterogeneous, parameter-rich tools, as well as the challenge of understanding and effectively leveraging their diverse low-level outputs (e.g., segmentation masks, depth maps) in high-level reasoning.
To address these challenges, we propose LAST, a unified framework for tool-augmented spatial reasoning.
LAST features an extensible interactive sandbox, termed LAST-Box, which abstracts heterogeneous tool invocations into atomic instructions and reusable spatial skills, returning multimodal hints (e.g., annotated images and textual descriptions) that can be directly consumed by LLMs.
We further design a three-stage progressive training strategy that guides models from understanding tool outputs to proficient and adaptive tool invocation.
Experiments on four datasets show that LAST-7B achieves around 20\% performance gains over its backbone and outperforms strong proprietary closed-source LLMs, substantially enhancing reasoning on complex spatial tasks.
\end{abstract}

\keywords{spatial reasoning, multimodal large language models, tool-augmented learning}

\maketitle


\section{Introduction}
Spatial reasoning ability~\cite{chen2024spatialvlm,yang2025vsibench} is a foundational capability for intelligent systems to perceive, plan, and execute tasks in the physical world. Whether in robot navigation~\cite{hong2021vln}, embodied AI interaction~\cite{liu2025embodiedsurvey}, or complex 3D scene understanding~\cite{li2025stibench}, models are required not only to recognize objects but also to accurately parse their geometric layouts and topological relationships. Recently, multimodal large language models (MLLMs) have made significant strides in general visual understanding and semantic reasoning~\cite{gupta2023visualprogramming,qwen2.5-VL}. However, these models often struggle with tasks involving precise spatial layouts, scale estimation, and fine-grained relationship reasoning. Existing research indicates that MLLMs frequently suffer from hallucinations regarding spatial relations~\cite{li2025viewspatial} or fail to comprehend complex perspective and depth information~\cite{chen2025sdvlm(msmu)}, which severely limits their potential in real-world physical scenarios and practical applications.

To mitigate these limitations, existing studies predominantly rely on data-driven scaling, i.e., fine-tuning models with increasing amounts of spatially related data~\cite{li2025spatialladder,chen2024spatialvlm,li2025viewspatial}. However, for spatial reasoning, these strategies often fail to enable models to truly internalize structured geometric priors and spatial constraints. In contrast, tool-augmented learning presents a more compelling alternative~\cite{yao2023react,qin2024tool1}. Given that many fundamental spatial perception skills, such as depth estimation~\cite{depthanythingv2} and precise object localization~\cite{groundingdino}, have already been effectively addressed by specialized vision models, a natural paradigm shift is to relieve MLLMs from relearning these mature capabilities and instead equip them with the ability to invoke expert tools. By integrating these highly optimized visual perception models, MLLMs can directly compensate for their perceptual deficiencies in complex spatial reasoning tasks.

\begin{figure*}[t]
    \centering

    \begin{subfigure}{0.48\linewidth}
        \centering
        \includegraphics[width=\linewidth]{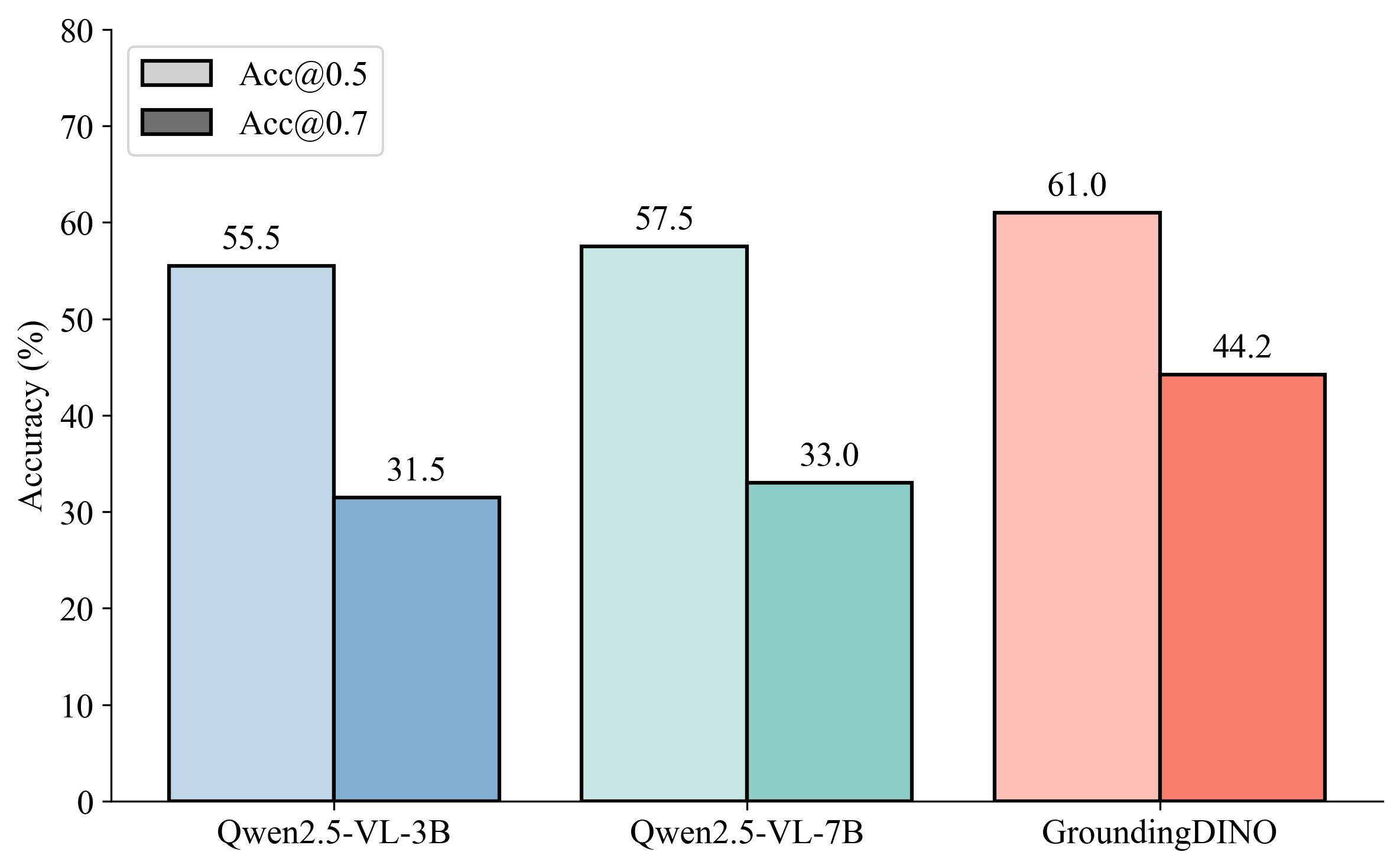}
        \Description{Bar chart comparing Acc@0.5 and Acc@0.7 localization accuracy of Qwen2.5-VL-3B, Qwen2.5-VL-7B, and GroundingDINO. GroundingDINO achieves the highest accuracy under both metrics.}
        \caption{Comparison of localization accuracy between MLLMs and specialized models. Light bars denote Acc@0.5 and dark bars denote Acc@0.7.}
        \label{fig:PA_Grounding}
    \end{subfigure}
    \hfill
    \begin{subfigure}{0.48\linewidth}
        \centering
        \includegraphics[width=\linewidth]{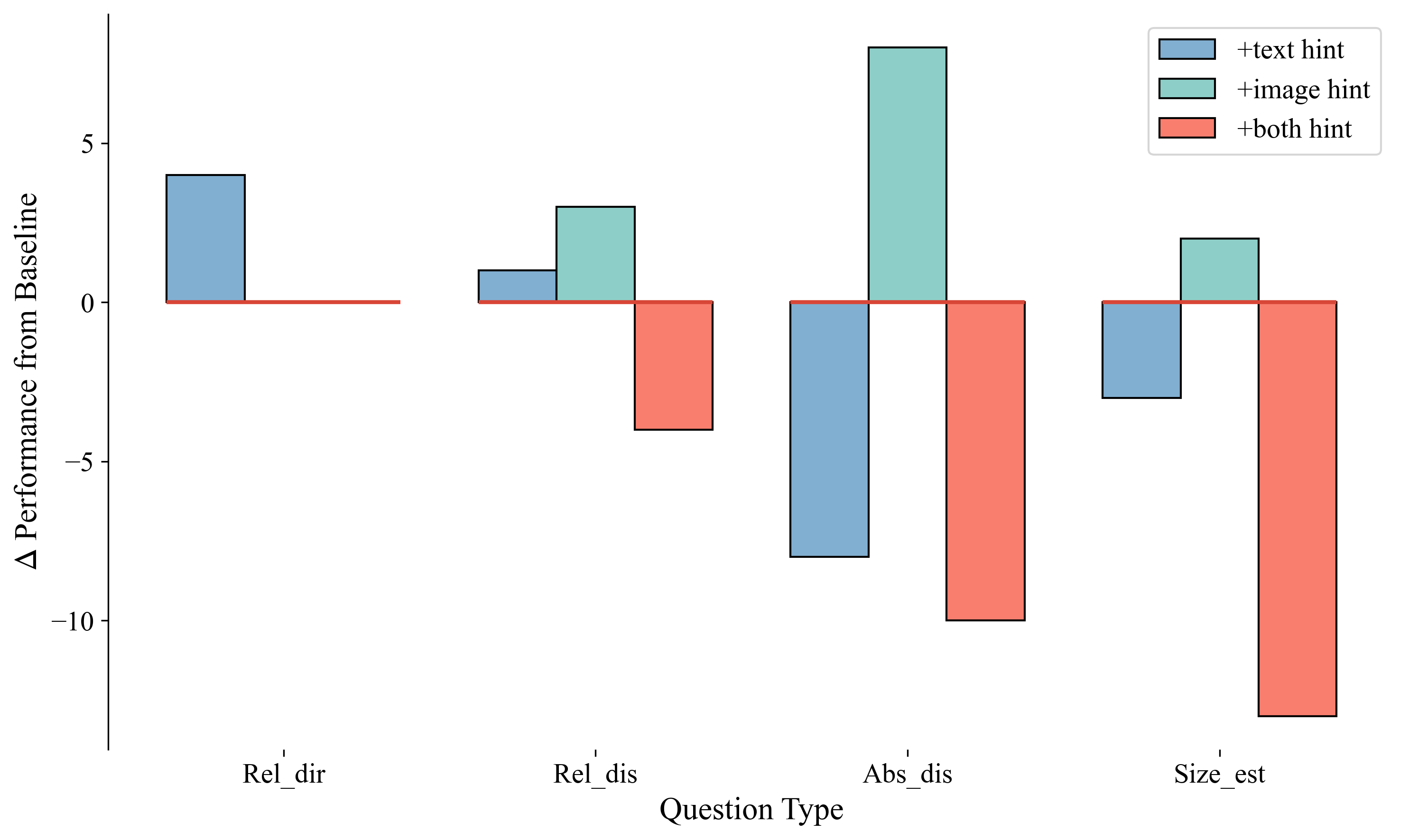}
        \Description{Grouped bar chart showing performance changes from baseline when using text hints, image hints, or both hints across four spatial task types.}
        \caption{Effects of different categories of tool hints across diverse tasks.}
        \label{fig:PA_hint}
    \end{subfigure}

    \caption{Visualization of preliminary experimental results for problem analysis.}
    \label{fig:pa_all}
\end{figure*}

However, in the domain of spatial reasoning, this paradigm encounters two distinctive challenges.
Firstly, spatial perception tools are heterogeneous and parameter-rich, requiring careful configuration at each step.
Errors in tool invocation can easily accumulate over long reasoning horizons, making robust multi-step tool usage by MLLMs difficult.
Secondly, perception tools produce diverse low-level outputs, such as segmentation masks or depth maps, which are not readily interpretable by MLLMs.
As a result, models struggle to understand these outputs and convert them into high-level spatial cues required for complex reasoning.

To address these challenges, we propose \method (\textbf{L}everaging Tools \textbf{as} Hin\textbf{t}s), a unified framework for tool-augmented spatial reasoning. \method introduces an extensible interactive sandbox, the \sandbox, which standardizes heterogeneous vision models into atomic instructions that can be composed into higher-level spatial skills.
These skills shorten the tool invocation horizon and encapsulate low-level tool outputs into model-interpretable textual and visual cues, enabling MLLMs to acquire perceptual capabilities in an intuitive and controlled manner.
To support effective tool interaction, we adopt a progressive curriculum-based training framework that guides models from understanding tool outputs to proficient tool invocation.
Experiments on four datasets show that \proposedmodel trained under this framework achieves around 20\% performance gains over the backbone and outperforms strong proprietary closed-source LLMs on complex spatial reasoning tasks.

Our main contributions are summarized as follows:
\begin{itemize}
\item \textbf{Atomic-to-Skill Tool Abstraction.}
We introduce a skill abstraction framework that composes atomic tool instructions into reusable spatial skills, substantially shortening long-horizon tool-calling trajectories.

\item \textbf{Progressive Tool-Aware Training.}
We propose a progressive training paradigm with a tool-aware warm-up stage, enabling models to better internalize tool semantics before learning complex instruction following.

\item \textbf{Open-Source 7B Model with Strong Performance.}
We release an open-source 7B multimodal model that integrates our sandbox and training paradigm, achieving state-of-the-art performance on multiple spatial reasoning benchmarks.
\end{itemize}

\section{Preliminary Analysis}




\textbf{\underline{Observation 1:} MLLMs still fall short of specialized models in fine-grained perception.}

To examine the differences between general-purpose MLLMs and specialized vision systems in foundational spatial reasoning, we conduct a comparative study on the grounding subtask.
We sample 200 instances from the SpatialLadder training set and benchmark the open-source Qwen2.5-VL series against the proprietary GroundingDINO.
As shown in Fig.~\ref{fig:PA_Grounding}, although scaling MLLMs leads to gradual improvements under both relaxed (Acc@0.5) and strict (Acc@0.7) metrics, their performance remains notably inferior to GroundingDINO, especially at higher localization precision.
Under the strict IoU threshold of 0.7, GroundingDINO achieves 44.2\% accuracy, whereas Qwen2.5-VL-7B attains only 33.0\%, revealing an 11.2\% performance gap.
Moreover, the limited gain from scaling the model from 3B to 7B (only +1.5\% in Acc@0.7) indicates that parameter scaling alone is insufficient for MLLMs to acquire precise geometric grounding.
These results highlight the persistent advantage of specialized vision models in fine-grained spatial perception and suggest the potential advantages of a tool-augmented paradigm that leverages their strengths.

\textbf{\underline{Observation 2:} Even with tool-based prompts, MLLMs struggle to effectively recognize and exploit such guidance across diverse spatial tasks.}

Given the distinct perceptual advantages of specialized models, we further investigated whether directly incorporating their outputs as ``hints'' could consistently enhance the spatial reasoning capabilities of MLLMs. To explore this, we sampled 100 instances from each subset of SPbench, covering four distinct task categories: two involving multiple-choice reasoning (Relative Direction and Relative Distance) and two requiring numerical estimation (Absolute Distance and Size Estimation). We utilized expert models, specifically GroundingDINO and SAM, to preprocess the images, formatting their predictions into either textual descriptions (text-hint) or visual overlays (image-hint). These tool-augmented inputs were then fed into the model alongside the original images for evaluation.

\begin{figure*}[t]
    \centering
    \includegraphics[width=0.9\linewidth]{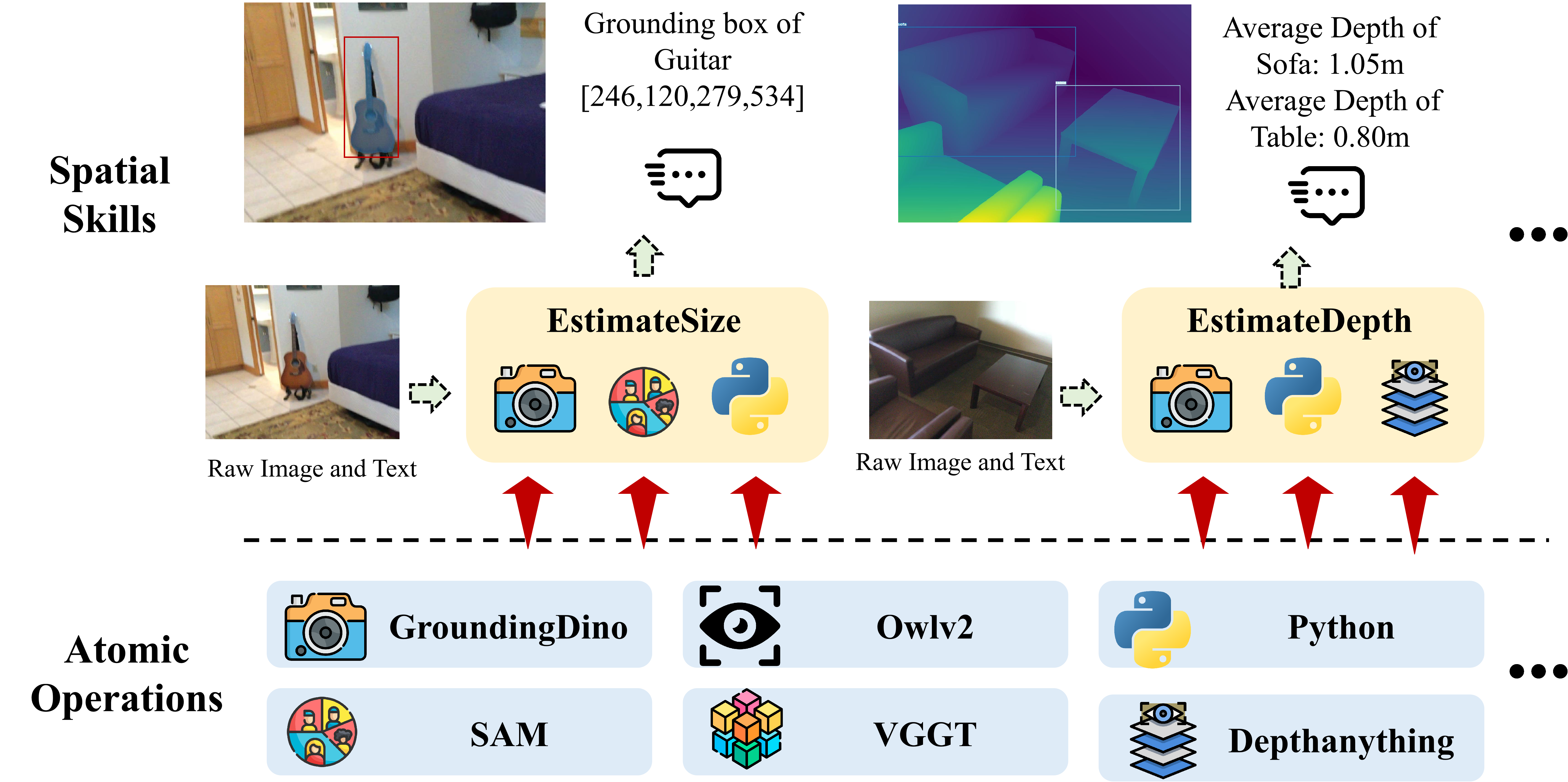}
    \Description{Architecture diagram of LAST-Box showing two layers: atomic operations (GroundingDINO, SAM, VGGT, DepthAnything, Python, Owlv2) at the bottom, and spatial skills (EstimateSize, EstimateDepth) at the top, with example inputs and outputs.}
    \caption{Overview of the proposed \sandbox}
    \label{fig:toolbox}
\end{figure*}

As illustrated in Fig.~\ref{fig:PA_hint}, the effectiveness of tool hints is highly dependent on the task nature. For tasks that are easily articulated in language, such as Relative Direction, textual hints provide a clear performance boost. Conversely, for tasks where textual description is inherently limited—such as metric estimation (Absolute Distance and Size Estimation)—visual image hints demonstrate a decisive advantage. Surprisingly, when both modalities are provided simultaneously (``both-hint''), the performance trend largely mirrors that of using text hints alone but consistently yields even lower scores. Instead of benefiting from richer information, the models appear overwhelmed, with the combined inputs causing severe performance degradation (e.g., dropping below -10 points in Size Estimation). This indicates that without specific training, MLLMs struggle to effectively process tool-augmented prompts, often treating the additional multimodal information as distractive noise rather than a reasoning aid.

\section{Methods}
We propose \method, which consists of a structured visual reasoning sandbox (\sandbox) and a progressive training strategy to enable efficient and controllable tool-augmented spatial reasoning for MLLMs.

\subsection{\sandbox}

We construct a comprehensive and extensible visual reasoning toolbox, denoted as \sandbox, designed to support controlled tool invocation and structured result parsing for MLLMs in complex spatial reasoning tasks.
Unlike prior work that directly exposes low-level model interfaces, \sandbox atomizes underlying model invocations into primitive skills and further composes them into higher-level spatial skills, allowing MLLMs to directly utilize well-formed skills and significantly shorten tool-calling trajectories, leading to more efficient and stable learning.
An overview of the \sandbox architecture is illustrated in Figure~\ref{fig:toolbox}.

\textbf{Atomic Operations.}
In \sandbox, we define a universal set of foundational modules $\mathcal{M} = \{m_1, m_2, \dots, m_K\}$. Each invocation of a module is treated as an \emph{atomic operation} $o \in \mathcal{O}$. Formally, an atomic operation is defined as a deterministic mapping:
\begin{equation}
    o: \mathcal{X}_o \to \mathcal{Y}_o
\end{equation}
where the input space $\mathcal{X} = \mathcal{I} \cup \mathcal{T}$ encompasses raw images $\mathcal{I}$ and textual prompts $\mathcal{T}$. The output space $\mathcal{Y}$ consists of structured symbolic representations, such as segmentation masks, bounding box coordinates, or depth matrices. 
Specifically, the atomic operations integrated into \sandbox comprise GroundingDINO~\cite{groundingdino} for object detection and text–region alignment, DepthAnything-v2~\cite{depthanythingv2} for monocular relative depth estimation, Segment Anything (SAM)~\cite{Kirillov_2023sam} for high-precision instance segmentation, VGGT~\cite{Wang_2025_vggt} for 3D reconstruction, as well as Python utilities that support numerical computation and rule-based reasoning.

\textbf{Spatial Skills.}
On top of atomic operations, we introduce Skills $s \in \mathcal{S}$ as an intermediate abstraction layer. A skill $s = \langle \{o_j\}_{j=1}^m, \mathcal{R} \rangle$ is a functional composition where $\mathcal{R}$ denotes the orchestration logic. While a skill shares the same input space $\mathcal{X}$ as atomic operations, its output space $\mathcal{V}$ is a high-level encapsulation of multimodal pairs:
\begin{equation}
    s: \mathcal{X} \to \mathcal{Y}_s, \quad \text{where } \mathcal{Y}_s = \{ \langle v_i, t_i \rangle \}_{i=1}^n
\end{equation}
Each element in $\mathcal{V}$ couples a visual output $v_i$ (visual output rendered via Python utilities, e.g., a cropped image patch) with its corresponding semantic text description $t_i$. For a given input $\mathbf{x} \in \mathcal{X}$, the skill execution is:
\begin{equation}
    s(\mathbf{x}) = \mathcal{R}(o_{i_1}, o_{i_2}, \dots, o_{i_m}; \mathbf{x})
\end{equation}

\begin{figure*}[t!]
    \centering
    \includegraphics[width=0.9\linewidth]{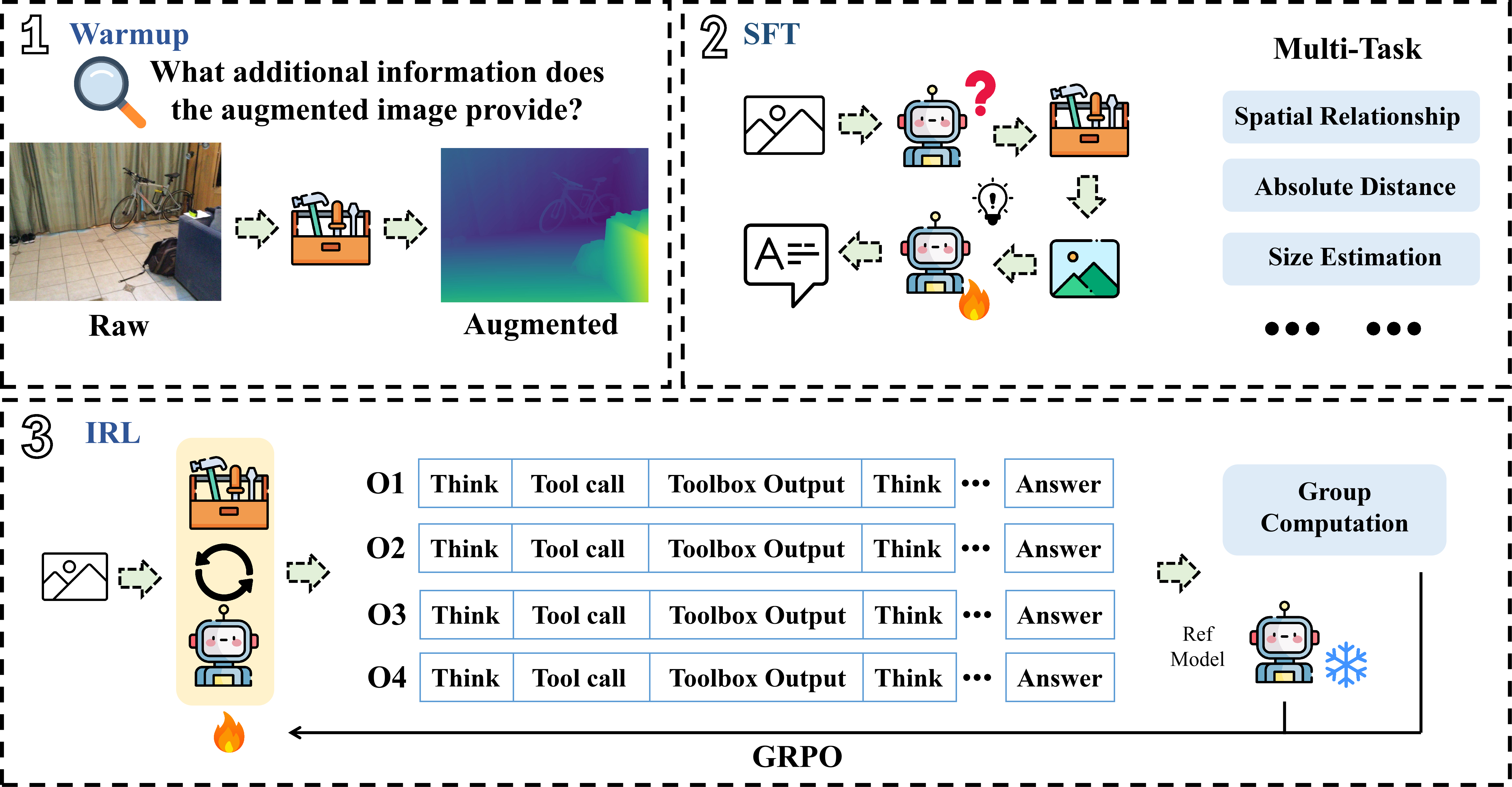}
    \Description{Diagram illustrating the three-stage progressive training strategy: Stage 1 Warmup with raw and augmented image pairs, Stage 2 SFT with multi-task spatial reasoning, and Stage 3 IRL with GRPO optimization over multiple rollouts.}
    \caption{Illustration of the progressive training strategy.}
    \label{fig:frame}
\end{figure*}

Where $R$ represents the combination rule of atomic operations. This design yields large-model-friendly interfaces where each $s: \mathcal{X}_s \to \mathcal{Y}_s$ provides a semantically explicit and functionally stable mapping, shielding the agent from the stochasticity and complexity of low-level model parameters.
For example:
\begin{itemize}
    \item \textbf{EstimateDepth} is implemented by combining GroundingDINO, DepthAnything-v2 and Python. When this Skill is invoked in \sandbox, the system returns a depth map overlaid with detection bounding boxes, together with the average relative depth of objects within each detected region.
    \item \textbf{EstimateSize} integrates GroundingDINO, SAM and Python. GroundingDINO first performs coarse localization, followed by precise segmentation via SAM, while Python provides an image-based visualization.
    The Skill outputs cropped image patches containing accurate instance segmentation masks, facilitating downstream analysis or external applications.
\end{itemize}

Based on the core spatial capability gaps identified in our preliminary analysis (Section~2), \sandbox currently encapsulates six skills targeting essential spatial reasoning needs: \textbf{SegmentObjects}, \textbf{EstimateDepth}, \textbf{EstimateSize}, \textbf{CountObjects}, \textbf{ZoomCrop}, and \textbf{Get3DPoint}. These skills collectively cover object localization, depth perception, metric measurement, instance counting, local detail inspection, and 3D coordinate estimation. The skill set is designed to be extensible---new skills can be integrated by defining the corresponding atomic operation compositions and I/O formats. Detailed API specifications for each skill are provided in Appendix~A.

\subsection{Progressive Training Strategy}
Inspired by prior work~\cite{li2025spatialladder}, we adopt a progressive tool-scheduling training strategy that progressively teaches the model how to invoke tools and how to reason over tool outputs and intermediate results in a shallow-to-deep manner.
Our training pipeline consists of three stages. An overall schematic diagram of the framework is shown in Figure~\ref{fig:frame}.

\textbf{Stage 1: Warm-up.}
The objective of the warm-up stage is to familiarize the model with the formats and semantics of tool outputs, particularly image-based intermediate results such as segmentation masks and depth maps.
We source 6,000 samples from the grounding subset of the SpatialLadder dataset. For each image, we utilize off-the-shelf vision tools to generate two types of augmented views: \textbf{mask maps} (highlighting specific objects) and \textbf{depth maps} (representing spatial geometry).
We construct QA pairs that explicitly ask the model to distinguish between these views and interpret their semantic content, with ground-truth answers generated by Qwen-VL-Plus. This step ensures that when the model invokes a tool in later stages, it can correctly comprehend the visual feedback (e.g., understanding that a depth map represents distance rather than texture).

\textbf{Stage 2: Supervised Fine-Tuning (SFT).}
In the second stage, we aim to establish a basic understanding of when and how to invoke tools.
We compile a dataset of approximately 16,000 samples, consisting of 6,000 single-image samples from the SpatialLadder~\cite{li2025spatialladder} training set and 10,000 samples from the MSMU~\cite{chen2025sdvlm(msmu)} dataset.
Using a state-of-the-art closed-source LLM (GPT-4o) as a teacher, we rewrite the original ground-truth labels into reasoning trajectories that follow a \textit{Think-Act-Observe-Reason} cycle, where the model decides to call a tool, analyzes the returned result, and derives the final answer.

To prevent the model from over-relying on tool outputs, we introduce a consistency-check mechanism during data construction.
For example, when a question asks for the spatial relationship between a refrigerator and a sofa, but GroundingDINO detects only one of the two objects, the tool's output is considered partially correct.
In such cases, the model is explicitly required to combine the original image with the incomplete tool output, rather than directly trusting the tool result.
Furthermore, in approximately 3,000 samples, we simulate tool failures (e.g., empty returns or execution errors). In these scenarios, the supervision signal encourages the model to abandon the tool path and revert to the original image for reasoning, thereby establishing a robust fallback mechanism.
We optimize the model by next-token prediction loss.

\begin{equation}
\mathcal{L}_{\mathrm{SFT}}(\theta)
= - \frac{1}{T} \sum_{t=1}^{T}
\log p_{\theta}\!\left(x_t \mid x_{<t}\right).
\end{equation}

\textbf{Stage 3: Interactive Reinforcement Learning (IRL).}
In the final stage, we further enhance the model’s capabilities in tool invocation decision-making and tool output interpretation through reinforcement learning.
We construct a dataset of 2,000 samples using the same source distribution as Stage 2. Unlike the SFT stage, these samples contain only the problem prompts without reasoning paths, forcing the model to explore and generate its own action trajectories during training.
We adopt GRPO~\cite{shao2024grpo} as the optimization algorithm and design a reward function composed of three components.
\begin{equation}
\begin{aligned}
&\mathcal{L}_{\mathrm{GRPO}}(\theta)
= \mathbb{E}_{q,\{o_i\}_{i=1}^G} \bigg[
\frac{1}{G} \sum_{i=1}^G \bigg(
\min \left(
r_i(\theta)\,\hat{A}_i,
\right. \\
&\quad \left.
\text{clip} \left(
r_i(\theta), 1-\epsilon, 1+\epsilon
\right)\hat{A}_i
\right)
- \beta \mathbb{D}_{\mathrm{KL}}(\pi_\theta \,\|\, \pi_{\mathrm{ref}})
\bigg)
\bigg]
\end{aligned}
\end{equation}

where  $r_i(\theta) = \frac{\pi_\theta(o_i \mid q)}{\pi_{\theta_{\mathrm{old}}}(o_i \mid q)}$ and the advantage $\hat{A}_i$ for each output $o_i$ in the group is computed as: $\hat{A}_i = \frac{r_i - \mu_r}{\sigma_r}$.

\textbf{Format reward}, which is used to enforce whether the model outputs conform to the expected structure.
\begin{equation*}
     r_{format} = \begin{cases} 
0 & \text{if } \forall t \in \mathcal{T}, \text{count}(t_{start}) == \text{count}(t_{end}) \\
-1 & \text{otherwise}
\end{cases}
\end{equation*}
where $\mathcal{T}$ denotes the set of tag types, and
$\mathrm{count}(t_{\mathrm{start}})$ and $\mathrm{count}(t_{\mathrm{end}})$
represent the numbers of opening and closing tags of type $t$, respectively (e.g., \texttt{<action></action>}).

\textbf{Correctness reward}, which measures the consistency between the predicted answer and the ground-truth label:
\begin{equation*}
    r_{correct} = \begin{cases} 
\mathbb{I}(\hat{y} = y) & \text{for discrete tasks} \\
\exp(-\alpha |\hat{y} - y|) & \text{for numerical tasks}
\end{cases}
\end{equation*}
Here, $\hat{y}$ and $y$ denote the predicted and ground-truth answers, respectively,
and $\alpha$ controls the sensitivity to numerical prediction errors.

\textbf{Tool reward.}
This reward evaluates whether the model invokes tools appropriately and whether the tool outputs are useful for task completion:
\begin{equation*}
r_{\mathrm{tool}}
= \mathbb{I}\!\left(c \in \mathcal{C}_{\mathrm{success}}\right) \mathbb{I}(\hat{y} = y)
\end{equation*}
where $c$ denotes the tool call issued by the model, $\mathcal{C}_{\mathrm{success}}$
is the set of successful tool calls, and
$\mathbb{I}(\hat{y} = y)$ represents the final result being correct.

The final reward is defined as a linear combination of these three components. The hyperparameter $\lambda$ balances the three reward terms.
\begin{equation*}
r_{all} =\lambda_1r_{format} + \lambda_2 r_{correct} + \lambda_3 r_{tool}
\end{equation*}

\begin{table*}[t]
\centering
\small
\setlength{\tabcolsep}{4pt} 
\renewcommand{\arraystretch}{1.2}

\definecolor{spbench_pink}{RGB}{253,238,238} 
\definecolor{group_gray}{RGB}{245,248,232}  

\caption{Comparison between our \proposedmodel and other baseline models. Numerical estimation tasks are highlighted with a pink background. \textbf{Bold} indicates the best performance in each category, while \underline{underline} denotes the second best.}
\label{tab:main_results}

\begin{tabular}{lcccccccccc}
\toprule
\multirow{2}{*}{Models}
& \multicolumn{3}{c}{CVbench}
& \multicolumn{4}{c}{SPBench-SI}
& \multirow{2}{*}{EmbSpatial}
& \multirow{2}{*}{MSMU}
& \multirow{2}{*}{Avg} \\
\cmidrule(lr){2-4} \cmidrule(lr){5-8}

& 2D-REL
& 3D-Depth
& 3D-Dist
& Rel\_Dir
& Rel\_dist
& Size\_Est
& Abs\_Dist
& 
& 
& \\
\midrule

\rowcolor{group_gray}
\multicolumn{11}{l}{\textbf{General Models}} \\

Qwen2.5-VL-3B
& 72.46 & 79.00 & 65.00
& 27.45 & 67.39 & \cellcolor{spbench_pink}18.61 & \cellcolor{spbench_pink}10.67
& 56.29 & \cellcolor{spbench_pink}8.67 & 45.06 \\

LLAVA-NEXT-8B
& 70.62 & 66.67 & 60.30
& 23.43 & 72.34 & \cellcolor{spbench_pink}27.63 & \cellcolor{spbench_pink}15.34
& 50.30 & \cellcolor{spbench_pink}10.63 & 44.14 \\

LLAVA-OneVision-8B
& 78.23 & 78.23 & 75.77
& 45.27 & 50.63 & \cellcolor{spbench_pink}35.70 & \cellcolor{spbench_pink}\underline{25.00}
& 56.20 & \cellcolor{spbench_pink}14.00 & 51.00 \\

InternVL3-8B
& 73.85 & 83.17 & 67.83
& 32.03 & 23.08 & \cellcolor{spbench_pink}24.62 & \cellcolor{spbench_pink}20.13
& 58.98 & \cellcolor{spbench_pink}13.44 & 44.13 \\

InternVL2.5-8B
& 70.63 & 78.15 & 71.33
& 30.42 & 40.29 & \cellcolor{spbench_pink}30.63 & \cellcolor{spbench_pink}18.56
& 50.82 & \cellcolor{spbench_pink}12.20 & 44.78 \\

\midrule
\rowcolor{group_gray}
\multicolumn{11}{l}{\textbf{Spatial VLMs}} \\

SpaceLLaVA-13B
& 62.03 & 68.23 & 63.16
& 40.46 & 55.37 & \cellcolor{spbench_pink}20.33 & \cellcolor{spbench_pink}16.67
& 67.40 & \cellcolor{spbench_pink}15.67 & 45.48 \\

Robopoint-13B
& 75.13 & 80.33 & 74.23
& 50.26 & 61.23 & \cellcolor{spbench_pink}25.83 & \cellcolor{spbench_pink}15.30
& 60.21 & \cellcolor{spbench_pink}17.28 & 51.09 \\

VILASR-7B
& 80.15 & \textbf{94.83} & \underline{88.33}
& \underline{64.05} & 53.85 & \cellcolor{spbench_pink}37.58 & \cellcolor{spbench_pink}4.70
& 69.81 & \cellcolor{spbench_pink}20.63 & 57.10 \\

\midrule
\rowcolor{group_gray}
\multicolumn{11}{l}{\textbf{Closed Source APIs}} \\

GPT-5.1
& 60.60 & 82.93 & 76.76
& 51.31 & 64.84 & \cellcolor{spbench_pink}46.87 & \cellcolor{spbench_pink}8.83
& 75.13 & \cellcolor{spbench_pink}18.69 & 54.00 \\

Gemini-3.0-flash
& 58.64 & 68.90 & 73.20
& 48.04 & 73.63 & \cellcolor{spbench_pink}50.54 & \cellcolor{spbench_pink}14.09
& 63.27 & \cellcolor{spbench_pink}25.07 & 52.82 \\

Gemini-2.5-pro
& 67.48 & 79.41 & 80.35
& 49.67 & 71.43 & \cellcolor{spbench_pink}\underline{52.27} & \cellcolor{spbench_pink}14.77
& \underline{77.27} & \cellcolor{spbench_pink}\underline{27.52} & \underline{57.80} \\

\midrule
\rowcolor{group_gray}
\multicolumn{11}{l}{\textbf{Ours}} \\

Qwen2.5-VL-7B (Backbone)
& \underline{81.15} & 80.00 & 73.67
& 39.22 & \underline{78.26} & \cellcolor{spbench_pink}41.99 & \cellcolor{spbench_pink}20.00
& 69.70 & \cellcolor{spbench_pink}14.20 & 55.35 \\

\textbf{\proposedmodel}
& \textbf{93.33} & \underline{89.54} & \textbf{90.64}
& \textbf{70.26} & \textbf{80.22} & \cellcolor{spbench_pink}\textbf{74.95} & \cellcolor{spbench_pink}\textbf{41.61}
& \textbf{81.00} & \cellcolor{spbench_pink}\textbf{41.17} & \textbf{73.64} \\

\midrule
\textit{Improvement}
& +12.18 & +9.54 & +16.97
& +31.04 & +1.96 & \cellcolor{spbench_pink}+32.96 & \cellcolor{spbench_pink}+21.61
& +11.30 & \cellcolor{spbench_pink}+26.97 & +18.28 \\

\bottomrule
\end{tabular}
\end{table*}

\section{Experiment}

\begin{figure*}[t]
    \centering
    \includegraphics[width=\linewidth]{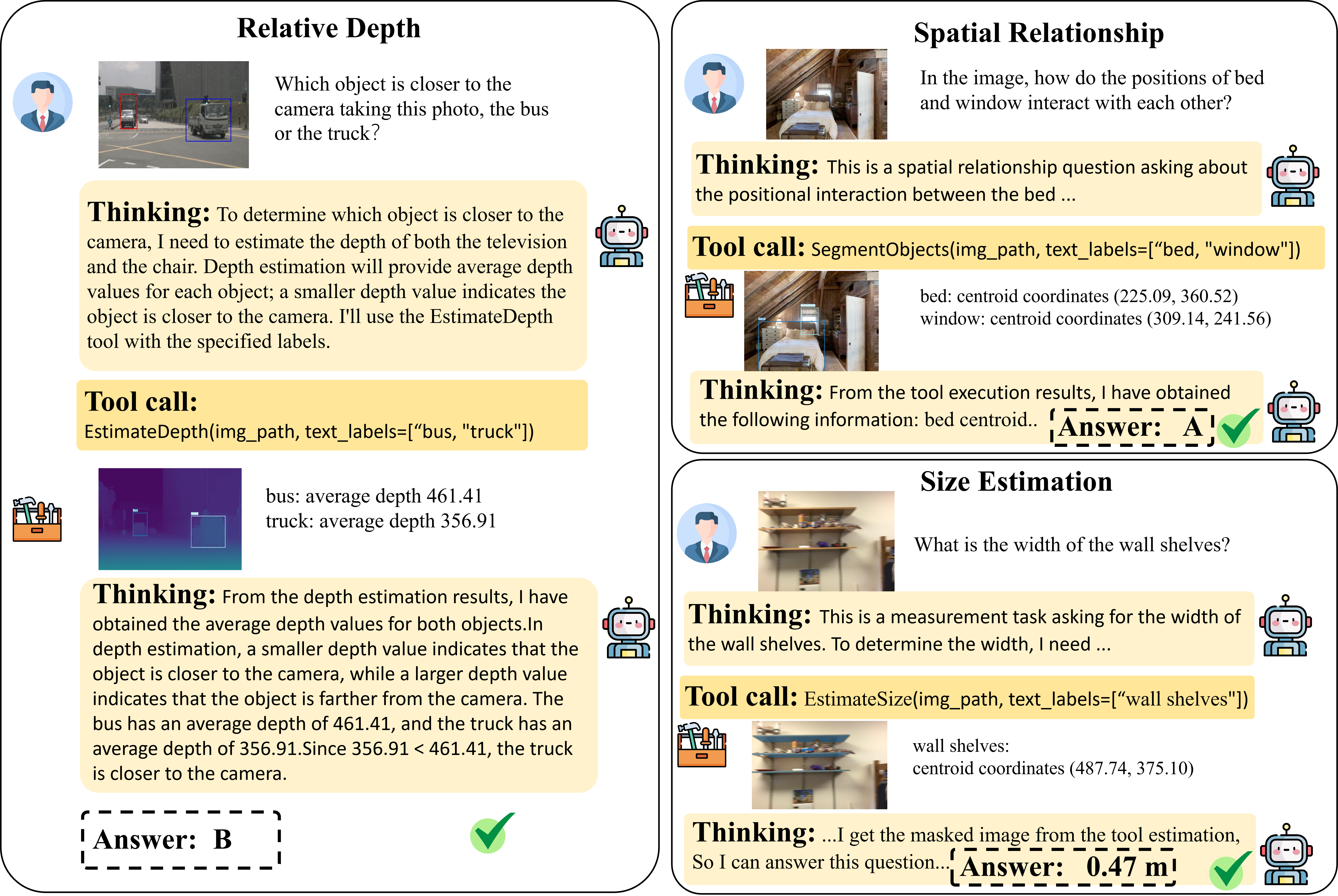}
    \Description{Four qualitative examples showing LAST-7B reasoning traces with tool calls for relative depth, spatial relationship, size estimation, and distance tasks, each with thinking steps, tool invocations, and final answers.}
    \caption{Representative experimental examples of \proposedmodel from CVBench, EmbSpatial, and MSMU.
}
    \label{fig:example}
\end{figure*}

\subsection{Setup}

\textbf{Datasets.}
We conduct experiments on four spatial reasoning benchmarks.
\textbf{CVBench}~\cite{tong2024cvbench} evaluates 2D visual recognition and 3D spatial perception across diverse visual tasks.
\textbf{EmbSpatial}~\cite{du2024embspatial} focuses on egocentric spatial understanding in embodied 3D environments.
\textbf{SPBench-SI}~\cite{li2025spatialladder} assesses single-image spatial reasoning including object localization, relative distance estimation, and spatial relation understanding.
\textbf{MSMU}~\cite{chen2025sdvlm(msmu)} targets quantitative spatial perception, requiring models to output precise metric values such as physical dimensions and absolute distances.
Further details of each benchmark are provided in Appendix~B.
The evaluated tasks fall into two main categories: multiple-choice questions and numerical fill-in-the-blank questions.
For evaluation metrics, multiple-choice questions are assessed using answer accuracy.
For scale and distance estimation tasks, a prediction is considered correct if the estimated value falls within an acceptable relative error margin $r$ with respect to the ground truth, i.e., $\text{ans} / \text{gt} \in [1 - r, 1 + r]$.
In our experiments, we set $r = 0.25$.

\textbf{Baselines.}
We mainly compare three categories of multi-modal large language models:
(1) General-purpose multimodal large language models (Qwen-2.5-vl-3B,7B~\cite{qwen2.5-VL}, LLava-Next-8B~\cite{zhang2024llavanextvideo},LLava-Onevison-8B~\cite{li2024llavaonevisioneasyvisualtask}, Internvl-2.5-8B~\cite{chen2024internvl} and Internvl-3-8B~\cite{zhu2025internvl3}),
(2) Specialized large models for spatial reasoning(SpaceLLava-13B~\cite{chen2024spatialvlm}, Robopoint-13B~\cite{yuan2024robopoint} and VILASR-7B~\cite{wu2025vilasr})
(3) Closed-source state-of-the-art multimodal large model APIs (GPT-5.1, Gemini-3.0-flash and Gemini-2.5-Pro).

\textbf{Implementation Details.}
We chose Qwen-2.5-VL-7B as the backbone. Both the supervised fine-tuning (SFT) and reinforcement learning (RL) stages are implemented using the \texttt{ms-swift}~\cite{zhao2025swift} framework.
For \textbf{SFT}, we perform full parameter fine-tuning with a global batch size of 64 (via gradient accumulation) and a learning rate of $1 \times 10^{-5}$ with a warmup ratio of 0.05 over 5 epochs. The maximum sequence length is set to 8,192 tokens.
For \textbf{RL}, we further align the model using GRPO over 2 epochs with a learning rate of $1 \times 10^{-7}$, employing a cosine decay scheduler (minimum learning rate ratio of 0.1) and a warmup ratio of 0.03. The reward weights for correctness, format, and tool usage are set to $1.0$, $0.3$, and $0.3$, respectively. The KL divergence coefficient ($\beta$) is set to 0.01. During rollout, we produce $G=4$ candidate completions per prompt with temperature 1.0, top-$p$ of 0.9, and top-$k$ of 50.



\subsection{Main Results}
In this section, we conduct a comprehensive analysis of our proposed \method and the trained model (\proposedmodel), obtaining the following experimental findings.





\textbf{\underline{Finding 1.} \proposedmodel substantially outperforms its backbone with an average gain of nearly 20\%, and even surpasses several closed-source API-based models.}
As evidenced by the comprehensive evaluation in Table 1, our proposed model achieves a new state-of-the-art (SOTA) performance, consistently outperforming all baseline methods across diverse spatial benchmarks. The comparison with the Qwen2.5-VL-7B backbone is particularly compelling; our method delivers a substantial average performance boost of 18.28 points (from 55.35 to 73.64), effectively transforming a strong general-purpose vision-language model into a high-precision spatial expert.
This enhancement is most pronounced in tasks requiring rigorous geometric and metric quantification rather than mere semantic description. Specifically, we observe remarkable gains in Size Estimation (+32.96), Relative Direction (+31.04), and MSMU (+26.97). These results suggest that our approach successfully bridges the long-standing ``metric gap'' in multimodal LLMs, enabling the model to transition from coarse relative perception to fine-grained spatial measurement.
Furthermore, the proposed model demonstrates superior scaling efficiency by significantly outperforming top-tier proprietary systems. Despite its 7B parameter size, it surpasses GPT-5.1 (Avg: 54.00) and Gemini-2.5-pro (Avg: 57.80) by margins of 19.64 and 15.84 points, respectively. Notably, our model maintains the highest or second-highest scores (indicated by bold and underline) in nearly every sub-metric, proving that specialized spatial alignment can yield reasoning capabilities that exceed even the most advanced closed-source large multimodal models.

\textbf{\underline{Finding 2.} The compressed skill abstraction in \sandbox effectively enhances the reasoning performance of closed-source models.}
To investigate the generalizability of our proposed interactive sandbox (\sandbox), specifically whether it can empower closed-source models that cannot be fine-tuned, we applied the framework to the Gemini model family and evaluated it across diverse spatial tasks. As shown in Table \ref{tab:cvbench_msmu_delta}, the results demonstrate that introducing external tool support yields significant performance gains in general.
Specifically, for the more capable Gemini-2.5-Pro, tool augmentation achieved consistent improvements across all metrics, delivering remarkable gains of 18.00\% and 14.90\% on CVBench's 2D relationship reasoning (2D-REL) and 3D distance estimation (3D-Dist) tasks, respectively. This evidence shows that when the base model possesses sufficient instruction-following and reasoning capabilities, it can effectively leverage the precise geometric signals provided by the sandbox to correct its intrinsic spatial perception biases.
Although we observed a performance drop (-8.06\%) for the smaller Gemini-3-Flash model on the 3D distance task, we attribute this degradation primarily to its limited reasoning capacity, which hampers robust generalization when multi-step spatial inference is required. In contrast, Gemini-3-Flash still exhibits substantial improvements of over 12\% on perception-dominant tasks such as 2D-REL and 3D-Depth, where the demands on long-horizon reasoning are less pronounced. Overall, these results indicate that our framework can also improve the spatial reasoning performance of closed-source commercial models.
\begin{table}[t]
\centering
\small
\setlength{\tabcolsep}{6pt}
\renewcommand{\arraystretch}{1.2}

\definecolor{deltagray}{RGB}{245,245,245}
\definecolor{improve}{RGB}{200,30,30}   
\definecolor{decline}{RGB}{0,140,0}     

\caption{Performance changes of closed-source models after integrating LAST-Box.}
\label{tab:cvbench_msmu_delta}

\resizebox{\linewidth}{!}{
\begin{tabular}{lcccc}
\toprule
\multirow{2}{*}{Models} 
& \multicolumn{3}{c}{CVBench} 
& \multirow{2}{*}{MSMU} \\
\cmidrule(lr){2-4}
 & 2D-REL & 3D-Depth & 3D-Dist &  \\
\midrule
Gemini-3-flash & 58.64 & 68.90 & 73.20 & 25.07 \\
+ LAST-Box     & 70.92 & 82.25 & 65.14 & 26.14 \\
\rowcolor{deltagray}
$\Delta$ (delta) 
& \textcolor{improve}{12.28$\uparrow$} 
& \textcolor{improve}{13.35$\uparrow$} 
& \textcolor{decline}{8.06$\downarrow$} 
& \textcolor{improve}{1.07$\uparrow$} \\
\midrule
Gemini-2.5-Pro & 67.48 & 79.41 & 80.35 & 27.52 \\
+ LAST-Box     & 85.48 & 90.32 & 95.25 & 30.06 \\
\rowcolor{deltagray}
$\Delta$ (delta) 
& \textcolor{improve}{18.00$\uparrow$} 
& \textcolor{improve}{10.91$\uparrow$} 
& \textcolor{improve}{14.90$\uparrow$} 
& \textcolor{improve}{2.54$\uparrow$} \\
\bottomrule
\end{tabular}
}
\end{table}

\definecolor{deltagray}{RGB}{245,245,245}
\definecolor{improve}{RGB}{200,30,30}   
\definecolor{decline}{RGB}{0,140,0}      

\begin{table}[t]
\centering
\small
\setlength{\tabcolsep}{6pt}
\renewcommand{\arraystretch}{1.2}
\caption{Tool success rate and accuracy under successful and unsuccessful tool usage across different datasets of \proposedmodel.}
\label{tab:tool_success_analysis}
\resizebox{\linewidth}{!}{
\begin{tabular}{lcccc}
\toprule
Dataset & Tool SR & Acc(w/uns) & Acc(w/suc) & \cellcolor{deltagray}$\Delta$Acc \\
\midrule
CVBench     & 88.67 & 87.65 & 91.51 & \cellcolor{deltagray}\textcolor{improve}{$\uparrow$ 3.86} \\
SPBench-SI  & 95.14 & 40.82 & 69.69 & \cellcolor{deltagray}\textcolor{improve}{$\uparrow$ 28.87} \\
EmbSpatial  & 99.67 & 60.00 & 80.74 & \cellcolor{deltagray}\textcolor{improve}{$\uparrow$ 20.74} \\
\bottomrule
\end{tabular}
}
\vspace{-2em}
\end{table}
\textbf{\underline{Finding 3.} The progressive training paradigm successfully teaches \proposedmodel to utilize tool outputs for improved reasoning.}
A critical question is whether \proposedmodel can effectively exploit information returned by tools. To investigate this, we analyze tool usage across datasets using the Tool Success Rate (Tool SR), and compare final accuracy under successful versus unsuccessful tool execution (Acc(w/suc) vs. Acc(w/uns)), as shown in Table~\ref{tab:tool_success_analysis}.
\proposedmodel achieves consistently high Tool SR across all evaluated benchmarks (88.67\%--99.67\%), indicating reliable tool invocation. More importantly, successful tool usage leads to substantial accuracy gains: on SPBench-SI and EmbSpatial, accuracy improves by approximately 29\% (69.69\% vs. 40.82\%) and 20\% (80.74\% vs. 60.00\%), respectively. Even on CVBench, where the backbone already performs well, successful tool execution still yields a meaningful +3.86\% improvement. These consistent gains confirm that the model meaningfully incorporates tool outputs into its reasoning process rather than invoking them superficially.
\begin{figure}[t]
    \centering
    \vspace{-0.5em}
    \includegraphics[width=\linewidth]{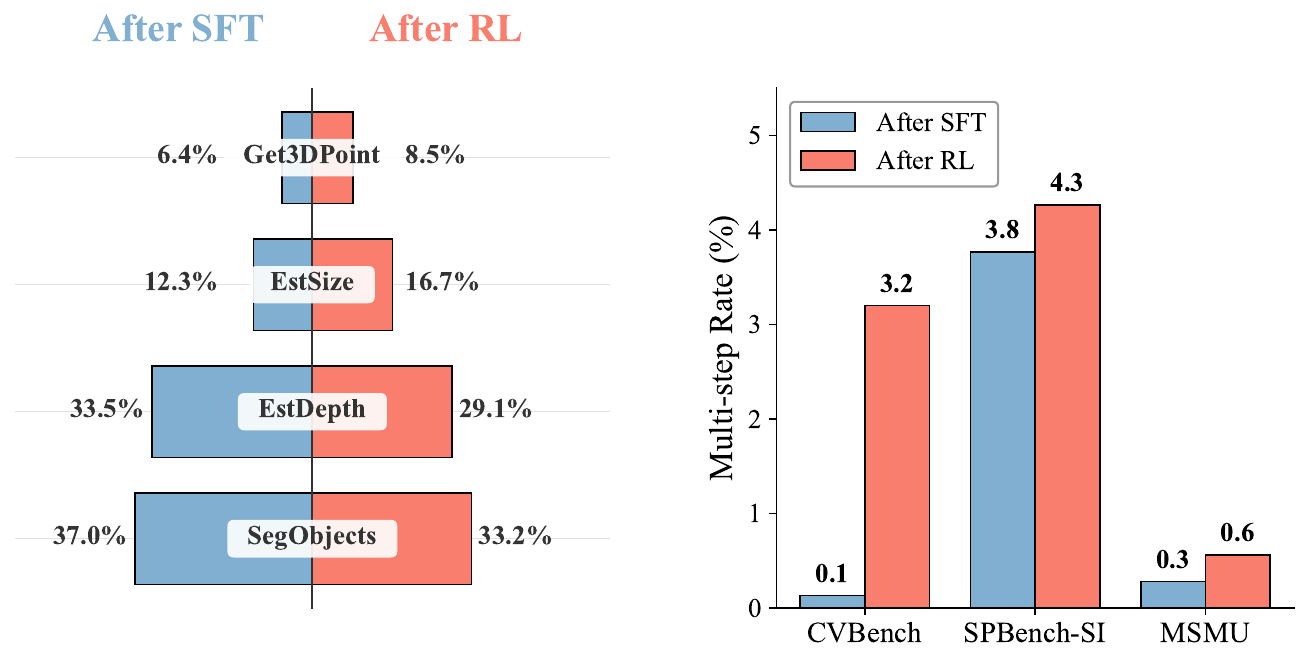}
    \Description{Two subfigures: (a) pie charts comparing tool usage distribution after SFT vs after RL, showing more balanced usage after RL; (b) bar chart showing multi-step tool invocation rates increase after RL across all benchmarks.}
    \vspace{-1em}
    \caption{Comparison of tool invocation behavior between After SFT (Stage 2) and After RL (Stage 3). (a) Top-4 tool usage distribution across benchmarks. (b) Multi-step invocation rate across benchmarks.}
    \label{fig:tool_rl}
    \vspace{-1em}
\end{figure}

Beyond tool success rates, we further investigate how RL training shapes tool invocation behavior compared to SFT alone (Figure~\ref{fig:tool_rl}). As shown in Fig.~\ref{fig:tool_rl}(a), RL training leads to a more balanced tool usage distribution across benchmarks, reducing the dominance of SegmentObjects (from 37.0\% to 33.2\%) while increasing the usage of EstimateSize (from 12.3\% to 16.7\%) and Get3DPoint (from 6.4\% to 8.5\%). Furthermore, as shown in Fig.~\ref{fig:tool_rl}(b), the RL model exhibits substantially higher multi-step tool invocation rates across all benchmarks, e.g., from 0.1\% to 3.2\% on CVBench. This indicates that RL training encourages the model to adaptively compose complementary tools for complex spatial reasoning tasks, rather than relying on a single tool call.

\subsection{Ablation Studies}

\definecolor{deltagray}{RGB}{245,245,245}
\definecolor{improve}{RGB}{200,30,30}   
\definecolor{decline}{RGB}{0,140,0}      

\begin{table}[t]
\centering
\caption{Ablation on training stages. Removing any stage leads to performance degradation.}
\label{tab:ablation_training}
\resizebox{\linewidth}{!}{
\begin{tabular}{lcccc}
\toprule
Settings & CVBench & SPBench-SI & EmbSpatial & MSMU \\
\midrule
Backbone     & 78.27 & 44.87 & 69.70 & 14.20 \\
w/o Stage 1 (Warmup)   & 89.63 & 64.87 & 79.63 & 39.23 \\
w/o Stage 2 (SFT)   & 79.30 & 48.26 & 65.45 & 20.63 \\
w/o Stage 3 (RL)   & 90.37 & 65.20 & 80.67 & 38.74 \\
\rowcolor{deltagray}
\proposedmodel & \textbf{91.17} & \textbf{66.76} & \textbf{81.00} & \textbf{41.17} \\
\bottomrule
\end{tabular}
}
\end{table}

\textbf{Ablation on Training Stages.}
To validate the effectiveness of each training stage, we conducted ablation studies as shown in Table \ref{tab:ablation_training}. The results demonstrate that the full proposed model achieves superior performance across all benchmarks. Specifically, removing Stage 2 (SFT) leads to the most significant performance degradation, highlighting its critical role in establishing the model's core capabilities. Furthermore, the absence of either Stage 1 (Warmup) or Stage 3 (RL) also results in consistent performance drops. This confirms that the initialization provided by Warmup and the refinement from RL are indispensable and complementary for achieving optimal performance.

\textbf{Ablation on Tool Hints.}
To study the role of tool hints, we compare three variants (Table~\ref{tab:ablation_tool}): \textit{No-tool SFT} trains with the same data but removes all tool outputs from the training trajectories; \textit{Text-only} and \textit{Image-only} use the fully trained model (\proposedmodel) but mask one hint modality during inference, retaining only the textual or visual output respectively.
The full model with both modalities consistently outperforms the no-tool variant across all benchmarks (e.g., +7.65 on CVBench, +2.77 on EmbSpatial). While single-modality hints can introduce noise on metric tasks like MSMU, fusing both modalities recovers and surpasses the baseline, confirming that text and image hints provide complementary spatial cues whose combination yields consistent improvements.

\section{Related Work}
\textbf{Spatial Reasoning.}
Spatial reasoning ability~\cite{liu2023visualspatial,chen2024spatialvlm,yang2025vsibench} is commonly defined as the capacity to interpret geometric relationships between objects and their surrounding environment~\cite{yang2025vsibench}. This capability is fundamental to robotic perception and is widely regarded as a core building block of embodied intelligence~\cite{liu2025embodiedsurvey} and autonomous driving~\cite{chen2024autodriving}.
Early spatial reasoning efforts largely focused on single-modality multi-step inference, whereas modern tasks predominantly target vision-based spatial reasoning with substantially higher complexity than conventional visual recognition problems (see Yang et al., 2025; VSI-Bench). Despite recent advances in visual-language models on basic perception, they continue to struggle with multi-step reasoning, precise scale estimation~\cite{chen2025sdvlm(msmu)}, and viewpoint transformation~\cite{li2025viewspatial}.
Benchmarking has also evolved—from static single-image settings (e.g., BLINK~\cite{fu2024blink}, CVBench~\cite{tong2024cvbench}) to dynamic multi-view reasoning scenarios (e.g., VSI-Bench~\cite{yang2025vsibench}, STI-Bench~\cite{li2025stibench}).
To bridge these gaps, the field has begun shifting from resource-intensive task-specific fine-tuning~\cite{cheng2024spatialrgpt} toward more generalizable approaches. Such approaches include progressive model training, integrating 3D-aware encodings~\cite{xu2024pointllm}, and modular systems that enable VLMs to invoke external computer vision tools to perform physical operations~\cite{han2025tigertoolintegratedgeometricreasoning,chen2025spacetools}. Recent research interests have begun to shift towards spatial reasoning in path planning~\cite{shang2026maptab,yang2026nesyroute}.

\begin{table}[t]
\centering
\caption{Ablation on tool hint modality. ``No-tool SFT'' trains with the same data but without any tool outputs. ``Image-only'' and ``Text-only'' restrict the hint modality during inference.}
\label{tab:ablation_tool}
\resizebox{\linewidth}{!}{
\begin{tabular}{lcccc}
\toprule
Settings & CVBench & SPBench-SI & EmbSpatial & MSMU \\
\midrule
Backbone           & 78.27 & 44.87 & 69.70 & 14.20 \\
No-tool SFT        & 83.52    & 63.81   & 78.23    & 39.73    \\
Text-only hint     & 86.77    & \textbf{66.90}    & 80.16    & 38.38    \\
Image-only hint    & 86.24    & 65.11    & 79.56    & 39.21    \\
\rowcolor{deltagray}
\proposedmodel (Full hint) & \textbf{91.17} & 66.76 & \textbf{81.00} & \textbf{41.17} \\
\bottomrule
\end{tabular}
}
\end{table}

\textbf{Tool-augmented Reasoning.}
A major research trend enhances LLMs by equipping them with external modules that supply complementary information. Typical examples integrate calculators~\cite{nakano2021toolcalculator1,zhang2025toolcalculator2}, code executors~\cite{gao2023pal,pan2023logiclm} and symbolic solvers~\cite{zhi2024neurodata,tian2025vcsearch,tian2025tabularmath,zhou2025lawgpt}, leveraging their reliability to handle complex reasoning beyond the native capacity of language models~\cite{tian2024crosel}.
In the multimodal setting~\cite{gupta2023visualprogramming,hu2024visualtool2}, tools are extended to visual operations such as cropping, masking, or adjusting image attributes~\cite{zhang2025thyme,zheng2025deepeyes}, sometimes coordinated through reinforcement learning for tool selection and sequencing~\cite{liu2025rlimagetool,zhou2025rlimagetool}.
Spatial reasoning marks another important direction. Text-based systems increasingly adopt logic engines or ASP solvers for multi-hop inference~\cite{yang2023spatialasp,wang2024nesyspatial}, while multimodal spatial reasoning~\cite{chen2025spacetools,han2025tigertoolintegratedgeometricreasoning} begins to exploit expert models for grounding and geometric analysis. However, existing approaches often depend on fixed prompting templates, limited interaction with tools, and mostly text-only outputs, restricting their flexibility and adaptability to dynamic environments.
In contrast, our framework offers a unified tool suite capable of producing heterogeneous outputs, and our inductive design enables systematic generalization across tasks and novel settings.


\section{Conclusion}
We propose \method, a unified tool-augmented framework that enhances the spatial reasoning capabilities of MLLMs through structured tool interaction, including the \sandbox that encapsulates heterogeneous vision backbone models into low-level atomic instructions and further composes them into short-horizon reusable spatial skills, effectively replacing long and error-prone tool-calling trajectories. These skills transform raw perceptual outputs into interpretable textual and visual hints that can be directly exploited by MLLMs. With a progressive curriculum training strategy, models learn to reliably understand and invoke spatial skills, achieving consistent improvements across four spatial reasoning benchmarks.
\bibliographystyle{ACM-Reference-Format}
\bibliography{example_paper}

\clearpage
\appendix



\section{\sandbox Details}
\setlist[description]{style=standard, labelsep=1em, leftmargin=2cm, font=\bfseries}
\setlist[itemize]{nosep, leftmargin=1.5em, labelsep=0.5em}

\subsection{Spatial Perception Skills}
\label{app:skill}

We provide a suite of atomic spatial perception tools integrated into our sandbox. Each tool exposes a standardized interface with explicit inputs and outputs, enabling seamless composition into higher-level spatial reasoning skills.

\paragraph{(1) \texttt{EstimateDepth}}
\begin{description}[leftmargin=2.5cm, style=nextline]
    \item[Description:] Runs Depth-Anything v2 to obtain a dense depth matrix and computes the average depth for specified objects. Produces both a depth visualization and numerical depth measurements.
    \item[Inputs:] 
        \begin{itemize}
            \item \texttt{img\_path} (\texttt{str}): Path to the input image.
            \item \texttt{text\_labels} (\texttt{list[str]}, optional): Object queries for depth aggregation, e.g., \texttt{["a person", "a frisbee"]}.
        \end{itemize}
    \item[Outputs:] 
        \begin{itemize}
            \item An image file path containing the depth visualization.
            \item A textual description reporting the average depth for each detected object (if \texttt{text\_labels} is provided).
        \end{itemize}
    \item[Example:]
\begin{verbatim}
EstimateDepth(img_path="image-0")
EstimateDepth(img_path="image-0",
  text_labels=["a person", "a frisbee"])
\end{verbatim}
\end{description}

\paragraph{(2) \texttt{EstimateSize}}
\begin{description}[leftmargin=2.5cm, style=nextline]
    \item[Description:] Performs open-vocabulary object detection using GroundingDINO and segments detected objects with SAM2. Produces dense masks and centroid coordinates for object size estimation.
    \item[Inputs:] 
        \begin{itemize}
            \item \texttt{img\_path} (\texttt{str}): Path to the input image.
            \item \texttt{text\_labels} (\texttt{list[str]}): Object queries to detect and segment.
            \item \texttt{threshold} (\texttt{float}, optional): Detection score threshold (default: 0.1).
        \end{itemize}
    \item[Outputs:] 
        \begin{itemize}
            \item An image file path visualizing segmentation masks.
            \item A textual description containing centroid coordinates for each detected object.
        \end{itemize}
    \item[Example:]
\begin{verbatim}
EstimateSize(img_path="image-0",
  text_labels=["lamp"])
EstimateSize(img_path="image-0",
  text_labels=["a person"],
  threshold=0.1)
\end{verbatim}
\end{description}

\paragraph{(3) \texttt{SegmentObjects}}
\begin{description}[leftmargin=2.5cm, style=nextline]
    \item[Description:] Detects and segments objects via GroundingDINO and SAM2 to support spatial relationship analysis.
    \item[Inputs:] 
        \begin{itemize}
            \item \texttt{img\_path} (\texttt{str}): Path to the input image.
            \item \texttt{text\_labels} (\texttt{list[str]}): Object queries to detect and segment.
            \item \texttt{threshold} (\texttt{float}, optional): Detection score threshold (default: 0.1).
        \end{itemize}
    \item[Outputs:] 
        \begin{itemize}
            \item An image file path showing segmentation results with bounding boxes.
            \item A textual description reporting centroid coordinates of detected objects.
        \end{itemize}
    \item[Example:]
\begin{verbatim}
SegmentObjects(img_path="image-0",
  text_labels=["a person", "a frisbee"])
\end{verbatim}
\end{description}

\paragraph{(4) \texttt{CountObjects}}
\begin{description}[leftmargin=2.5cm, style=nextline]
    \item[Description:] Uses open-vocabulary detection with GroundingDINO to count multiple instances of queried objects and report their spatial locations.
    \item[Inputs:] 
        \begin{itemize}
            \item \texttt{img\_path} (\texttt{str}): Path to the input image.
            \item \texttt{text\_labels} (\texttt{list[str]}): Object categories to detect and count.
            \item \texttt{threshold} (\texttt{float}, optional): Detection score threshold (default: 0.1).
        \end{itemize}
    \item[Outputs:] 
        \begin{itemize}
            \item An image file path with bounding boxes for all detected instances.
            \item A textual description reporting the count and centroid coordinates for each object type.
        \end{itemize}
    \item[Example:]
\begin{verbatim}
CountObjects(img_path="image-0",
  text_labels=["table"])
\end{verbatim}
\end{description}

\paragraph{(5) \texttt{ZoomCrop}}
\begin{description}[leftmargin=2.5cm, style=nextline]
    \item[Description:] Crops and optionally zooms into a region of interest (ROI) specified by a bounding box or by center and size.
    \item[Inputs:] 
        \begin{itemize}
            \item \texttt{img\_path} (\texttt{str}): Path to the input image.
            \item \texttt{box} (\texttt{list[float]}, optional): Bounding box \texttt{[x1, y1, x2, y2]} in pixels.
            \item \texttt{center} (\texttt{list[float]}, optional): ROI center \texttt{[cx, cy]} in pixels.
            \item \texttt{zoom\_factor} (\texttt{float}, optional): Zoom factor (default: 1.0).
        \end{itemize}
    \item[Outputs:] 
        \begin{itemize}
            \item An image file path of the cropped region.
            \item A textual description containing crop metadata.
        \end{itemize}
    \item[Example:]
\begin{verbatim}
ZoomCrop(img_path="image-0",
  box=[100, 200, 300, 400])
\end{verbatim}
\end{description}

\paragraph{(6) \texttt{Get3DPoint}}
\begin{description}[leftmargin=2.5cm, style=nextline]
    \item[Description:] Combines GroundingDINO for object detection with VGGT (Visual Geometry Grounded Transformer) to estimate the 3D coordinates of objects relative to the camera coordinate system.
    \item[Inputs:] 
        \begin{itemize}
            \item \texttt{img\_path} (\texttt{str}): Path to the input image.
            \item \texttt{text\_labels} (\texttt{list[str]}): Object queries to locate in 3D space.
        \end{itemize}
    \item[Outputs:] 
        \begin{itemize}
            \item An image file path showing 3D visualization.
            \item A textual description reporting \texttt{[X, Y, Z]} coordinates.
        \end{itemize}
    \item[Example:]
\begin{verbatim}
Get3DPoint(img_path="image-0",
  text_labels=["cup"])
\end{verbatim}
\end{description}

\section{Dataset Details}
\label{app:dataset}
\subsection{Evaluation Datasets}
\noindent \textbf{CVBench}~\cite{tong2024cvbench} is a vision-centric benchmark introduced in the Cambrian-1 study. It provides a comprehensive evaluation of Multimodal LLMs across diverse visual tasks, specifically assessing fundamental 2D visual recognition and 3D spatial perception capabilities to ensure robust visual grounding.

\noindent \textbf{EmbSpatial}~\cite{du2024embspatial} (EmbSpatial-Bench) focuses on spatial understanding within embodied AI scenarios. Unlike general visual benchmarks, it evaluates the model's ability to interpret spatial relationships and constraints from an \textit{egocentric} perspective in 3D environments, which is critical for embodied agents.

\noindent \textbf{SPBench-SI}~\cite{li2025spatialladder} is a subset of the Spatial Ladder evaluation framework designed for single-image spatial reasoning. It assesses the model's proficiency in tasks such as object localization, relative distance estimation, and spatial relation understanding within static 2D images, serving as a foundational test for spatial intelligence.

\noindent \textbf{MSMU}~\cite{chen2025sdvlm(msmu)} (Massive Spatial Measuring and Understanding) is a large-scale dataset specifically curated for \textit{quantitative} spatial perception. Distinct from qualitative VQA tasks, MSMU requires the model to output precise metric values (e.g., physical dimensions or absolute distances), thereby rigorously testing the model's capability in numerical spatial reasoning.

\section{Prompts}
\label{app:prompts}
\begin{figure*}[thb]
    \centering
    \begin{definedbox}[label=sys_prompt_tools]{System Prompt for Tool-Integrated Spatial Reasoning}
        \small

        \textbf{Role \& Goal:} \\
        You are a helpful assistant specialized in \textbf{spatial reasoning tasks}. Your goal is to solve spatial reasoning problems by analyzing images and using available tools. Each tool processes the image and returns an augmented view or numerical data to assist your reasoning.

        \vspace{0.4em} \hrulefill \vspace{0.4em}

        \textbf{Available Tools:}
        \begin{itemize}[leftmargin=1.5em, itemsep=0.1em, topsep=0.2em]
            \item \texttt{EstimateDepth(img\_path, text\_labels)}: Estimate depth information for objects. Returns a depth map and average depth values.

            \item \texttt{EstimateSize(img\_path, text\_labels)}: Estimate the size of objects by generating segmentation masks. Returns a mask image and centroid coordinates.

            \item \texttt{SegmentObjects(img\_path, text\_labels)}: Segment objects to obtain their masks and boundaries. Returns a segmentation image with bounding boxes.

            \item \texttt{CountObjects(img\_path, text\_labels)}: Detect and count multiple instances of specified objects. Returns a bounding box image and count statistics.

            \item \texttt{ZoomCrop(img\_path, box)}: Crop and zoom into a specific region of interest defined by a bounding box. Returns the cropped image.

            \item \texttt{Get3DPoint(img\_path, text\_labels)}: Estimate the 3D coordinates \texttt{[X, Y, Z]} of objects relative to the camera. Returns a 3D visualization and coordinates.
        \end{itemize}

        \vspace{0.4em} \hrulefill \vspace{0.4em}

        \textbf{Format Requirements:} \\
        Please carefully analyze the given image and question, then use the appropriate tools to solve the spatial reasoning problem step by step.
        \begin{itemize}[leftmargin=1.5em, itemsep=0.1em, topsep=0.2em]
            \item Your thinking process should be wrapped in \texttt{<analy></analy>} tags.
            \item Tool calls should be wrapped in \texttt{<action></action>} tags.
            \item Your final answer should be wrapped in \texttt{<ans></ans>} tags.
        \end{itemize}

    \end{definedbox}
    \caption{The system prompt designed to guide the model in selecting and executing visual tools.}
\end{figure*}

\end{document}